# Identification of Tree Species in Japanese Forests based on Aerial Photography and Deep Learning

Sarah Kentsch[1], Savvas Karatsiolis[2], Andreas Kamilaris[2,3], Luca Tomhave[1,4] and Maximo Larry Lopez Caceres[1]

[1] United Graduate School of Agricultural Sciences (UGAS)
Faculty of Agriculture, Yamagata University, Tsuruoka, Japan
[2] Research Centre on Interactive Media, Smart Systems and Emerging Technologies (RISE), Nicosia, Cyprus
[3] Department of Computer Science, University of Twente, The Netherlands
[4] Leibniz University Hannover, Germany
sarahkentsch@gmail.com, karatsioliss@cytanet.com.cy,
a.kamilaris@rise.org.cy, luca.tomhave@kabelmail.de,
larry@tds1.tr.yamagata-u.ac.jp

**Abstract.** Natural forests are complex ecosystems whose tree species distribution and their ecosystem functions are still not well understood. Sustainable management of these forests is of high importance because of their significant role in climate regulation, biodiversity, soil erosion and disaster prevention among many other ecosystem services they provide. In Japan particularly, natural forests are mainly located in steep mountains, hence the use of aerial imagery in combination with computer vision are important modern tools that can be applied to forest research. Thus, this study constitutes a preliminary research in this field, aiming at classifying tree species in Japanese mixed forests using UAV images and deep learning in two different mixed forest types: a black pine (*Pinus thunbergii*)-black locust (*Robinia pseudoacacia*) and a larch (Larix kaempferi)-oak (Quercus mongolica) mixed forest. Our results indicate that it is possible to identify black locust trees with 62.6 % True Positives (TP) and 98.1% True Negatives (TN), while lower precision was reached for larch trees (37.4% TP and 97.7% TN).

## 1 Introduction

Natural mixed forests are known as complex ecosystems with high resilience, high biodiversity, productivity and their carbon sink capacity. They play a role in the exchange of water carbon and nutrients within the soil-forest-atmosphere continuum. Under the present climate change conditions, high $CO_2$ emissions and degrading forest areas, made it essential to quantify the role of natural mixed forests on ameliorating the negative impact of anthropogenic emissions on climate change. Furthermore, the preservation of biodiversity, the physiological tolerances of species and effects of plant stress (due to droughts, pests and invasion) have to be considered. In particular, forests provide wood and non-wood resources, maintain soil fertility, regulate climate and preserve water supplies [1], [2]. Several studies proposed that a sound monitoring, stand inventories, quantification of tree species and ecosystem services are necessary to ensure their sustainability [3], [4], [5].

Forests in Japan occupy nearly 70 % of the total territory. Two-thirds of the forests are located in mountainous areas and half of the total forest area made of timber plantations [6]. Forest plantations have a long history of clear-cuts followed by reforestation. Those planted forests with their simple structure of trees and lower biodiversity [7] have replaced natural forest areas, leading to a decrease in tree species diversity. Recent efforts to restore natural forests, as a result of climate change, have influenced Japan's point of view [8]. Recreation and protection are now the main drivers for forest management efforts [6]. However, natural forests have not fallen into adequate management strategies [8]. Furthermore, natural mixed forests have not been adequately studied, while most of the existing relevant case studies were only carried out for small forest patches [9]. Thus, tree species' composition and diversity as well as their distribution and interaction within the forest ecosystem need to be studied. ,The first step is to develop a monitoring system that allows a rapid and reliable method to survey this type of forest and additionally to be able to identify tree species. The use of aerial imagery in combination with deep learning approaches are essential tools and techniques that can build the bases for the improvement of monitoring methodologies in forestry research [10], [11], [12], [13]. We used UAV (Unmanned Aerial Vehicles) to capture images of the forests and then trained a deep learning network to identify tree species in two different forests types. Specifically, we used one dataset to identify invasive black locust trees in coastal black pine forests from drone images and a second dataset to identify conifer trees. Since this work is the first attempt of our research team to combine forestry and AI, the main purpose is to evaluate the results of automatic tree species' identification.

## 2  Related Work

Previous studies used aerial images gathered by satellites, airplanes and UAVs to identify trees species in forests. The work in [14] classified tree species in a mixed forest by using high-resolution IKONOS data. Twenty-one species were classified by using panchromatic and multi-spectral bands. After pre-processing the images, 50 pixels per species were extracted and a Turkey´s multiple comparison test was applied. Finally, maximum likelihood classifiers were performed for reaching accuracies of 62%. According to [14] broadleaf trees are more difficult to classify than conifers.

In a comparison study of tree identification using IKONOS and WorldView-2 (WV2) images with a resolution of 1-4 m, overall accuracies of 57% were reported for 7 tree species and 15 selected features [15]. In this study, linear discriminant analysis (DLA) and decision tree classifier (CART) were used.

Dalponte et al. [16] used both high-resolution airborne hyperspectral images and satellite images, each in combination with LIDAR data to understand the classification potential by using different datasets. The data was pre-processed for normalization and generalization, as well as feature selection for LIDAR data. Support Vector Machines (SVM) and random forest classifiers (RFC) were used for the classifications. Different classes were tested, ranging from single tree species to macro-classes reaching kappa accuracies of 76.5-93.2%, concluding that hyperspectral data resulted in highest accuracies while SVM outperformed RFC. Torresan et al. [17] reported that 14% of UAV-related studies focused on tree species classification. One such study was carried out in [18], classifying tree species by using RGB and hyperspectral images of a boreal forest. In total, 11 orthomosaics and DSMs (Digital Surface Models) were used, as well as reference data. Spectral and 3D point cloud information was used to classify trees. RFC and multilayer perceptrons performed the classification with accuracies of around 95% for four different tree species. Moreover, the work in [19] focused on the best time-window to gather images by UAVs. Their primary aim was to effectively classify tree species using a multi-temporal dataset. The data was gathered in a broadleaf forest, composed by 577 tree species that were divided into 5 groups. Orthomosaics were used to analyze the spectral response, the characteristics and differences of tree classes. The pixel intensity was used to run RFC. Misclassifications of 15.9 and 36% were observed by using one dataset of one season only. The error was decreased to 8.8% when using multitemporal datasets. A method to classify tree species in a mixed forest dominated by pine trees using high-resolution RGB images collected by three-years was proposed in [20]. Orthomosaics and DSMs were used to delineate tree crowns in a first step by using local maxima filtering the watershed algorithm. The extracted tree crowns were used to train a Convolutional Neural Network (CNN). Therefore, two approaches were conducted: one using one orthomosaic and another using three orthomosaics for training and/or testing. Classification accuracies between 51-80% were recorded.

## 3 Methodology

### 3.1 Problem description

The first study site, where the first dataset was created, is located in the Yamagata prefecture on the Japanese northwest coast (38°49'14"N, 139°47'47"E). The coastal forest is a black pine (*Pinus thunbergii*) plantation (Fig. 2A) with high tolerances against acidity, alkalinity and salty soils and drought conditions. This forest was planted in order to protect the surrounding area from strong winds and sand movement. Since the early 1990's, this forest has been invaded by black locust trees (*Robinia pseudoacacia*) (Fig. 2B), a fast-growing species establishing in gaps of the black pine forest. Black locust species is known for its rapid invasion and high biomass production with a high impact on the structure and function of tree communities [21]. However, the exact influences on the functions of the coastal forest are unknown. Generally, invasive tree species have a high impact on the structures, properties and functions of natural ecosystems [22], [23]. Therefore, it is necessary as a first step to detect and identify black locust trees in order to provide information about their distribution and density, as a second step, to understand the structure and nutritional impact of this invasion on the black pine properties as a windbreak and growth. Information about these parameters will offer essential insights for a sound management of this type of forest. Additionally, this information offers the possibility to quantify the effects of invasive species as it turns the monoculture into a mixed forest, which is supposed to be less affected by diseases and infections [24]. The second site, which provides the second dataset, is located in the Yamagata University Research Forest (YURF) on the Japanese main island Honshu. It is located in the northern part of the Asahi Mountains. The research forest covers an area of 753 ha.

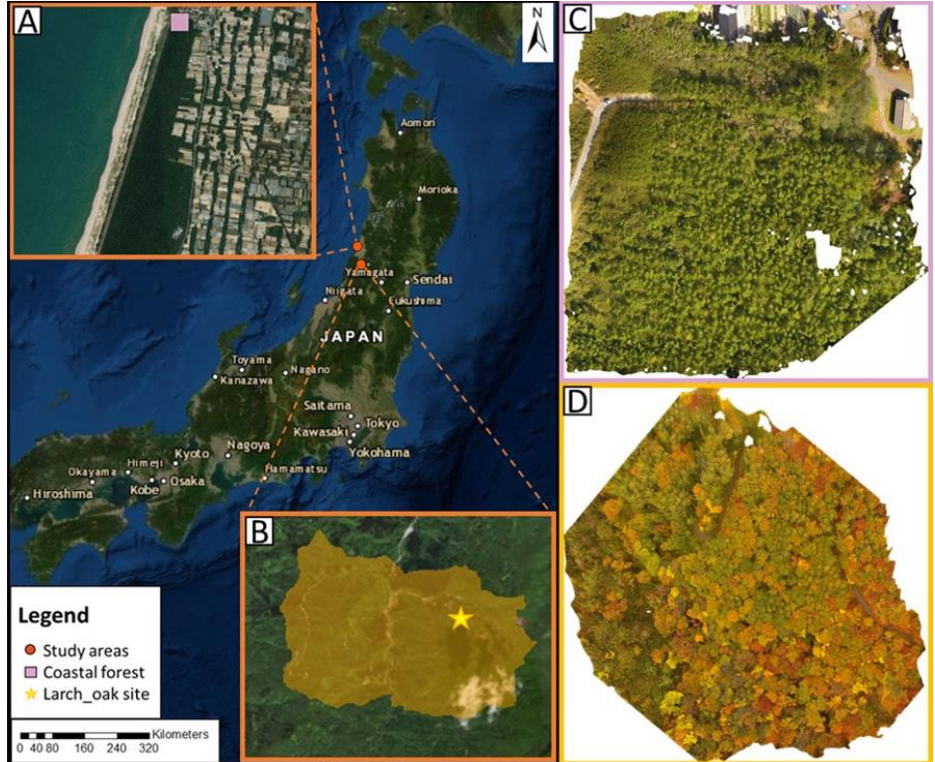

**Fig. 1.** The main map showing the location of the study areas in Japan, where data was acquired: A) Coastal forest near Sakata city. B) YURF (orange area), located south of Tsuruoka city in the Asahi mountains. C) Orthomosaic of the coastal forest, where images were collected to identify black locust trees that are invasive in this area. D) Orthomosaic of the larch oak site, where the minor tree species larch should be identified.

The forest is characterized by steep slopes (30 to 44 degrees) within a range of altitudes between 250 and 850 m and it is crossed almost in half by the Wasada River. The area is composed of a mixed natural forest, as well as deciduous broadleaf and coniferous trees [25]. Our study sites are located in a slope, as shown in Fig. 1. The second site is mainly composed of a mixture of larch (*Larix kaempferi*) trees and oak (*Quercus mongolica*) and in a minor proportion beech (*Fagus crenata*) (Fig. 2). Such as the YURF, the large majority of Japanese forests is located in steep mountains and is characteristic by a complex ecology. Thus, the conduction of field surveys is very much limited, primarily due to the lack of accessibility and the necessity of man-power. Therefore, developing an automatic methodology capable of identifying tree species in mixed forests is crucial for the evaluation of these forest ecosystems.

### 3.2 Data collection

Data acquisition was carried out using a DJI Phantom 4 drone. The drone is equipped with a 12-megapixel camera which produces high resolution geo-referenced images.

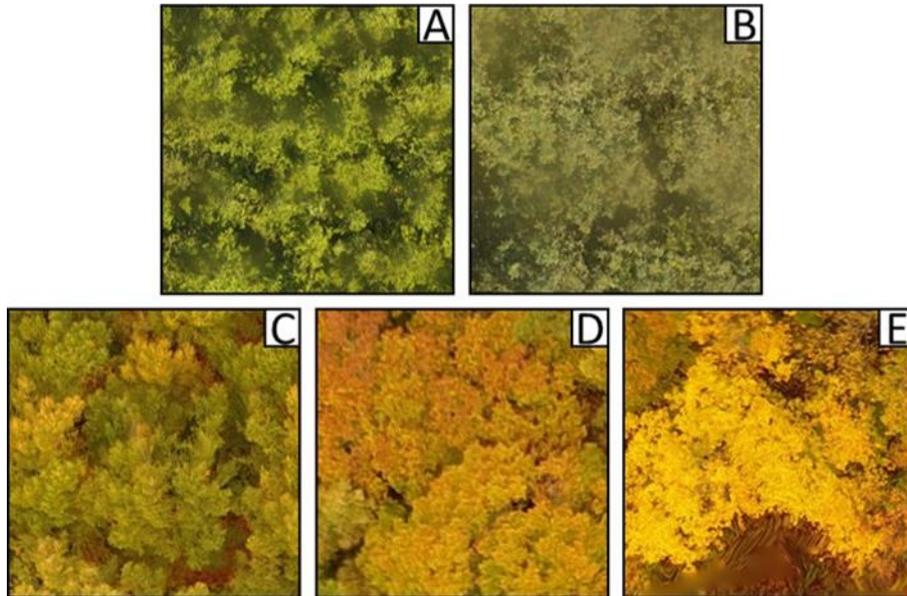

**Fig. 2.** Examples of tree species in our study areas, as shown from aerial photos. A) Pine trees and B) black locust represent trees of the coastal forest. The trees look similar to each other with small differences in the colour (due to the colour change of black locust trees), which increases the difficulty to identify them. C) larch, D) oak and E) beech represent trees of the YURF mixed forest. As can be seen in the images, autumn images were chosen since the trees show different coloured leaves which increase the potential to classify them.

The flights were performed using the autonomous mode, which standardizes the acquisition protocol. The coastal forest was photographed in July and October 2019, but only the analysis of images from October are presented in this study. Images from July failed to be automatically recognized (i.e. in a previous analysis we performed) since differences between tree species were too low. During the flight, around 1,100 images were collected with an overlap of 90%. The flight altitude was 30 m and the covered area was 2.7 ha. Since we already faced difficulties in identifying trees in the first case study, we chose from the beginning autumn images for the second one. Manual annotations were only available for this season since the forest expert experienced the same limitations in accurately identifying tree species in the summer orthomosaics. For the larch-oak site, an additional flight was performed at the end of October 2019, capturing 202 images with a set overlap of 93% covering an area of 3.2 ha. The collected images were processed with Agisoft Metashape [26], in order to generate the orthomosaics. The autumn images of the coastal forest were not aligned well and the orthomosaics had several blank spots. The resolution of the orthomosaics was approximately 1.1 cm/pix for the coastal forest sites and 3.5 cm/pix for the larch-oak sites.

### 3.3 Data pre-processing

The two datasets were annotated by forest experts, knowing the study area well. The annotations were done using the image editing software Gimp. Areas of each tree species were colored black and stored in different layers. Those layers were used as ground truth data. Due to the different visual characteristic of larch trees, the accuracy of that layer was assumed as high. More difficulties appeared during the annotation process of the coastal forest site due to the comparably small canopy area of the black locust and its mix with other broadleaved species. Furthermore, since a new approach for tree classification was used by the experts, a certain misclassification cannot be ruled out. Nonetheless, a high accuracy of annotations can be assumed for both sites.

Image segmentation aims to partition an image into semantically related segments so that each pixel in the image is assigned to a group of coherent pixels. Consequently, this enables image analysis focusing on its elements. The specific tree identification problem differs from general-case image segmentation in the sense that each image pixel represents a larger physical space ($\approx 1m^2$) and does not necessarily belong to a single component of a larger entity. Roughly, each pixel in a species map corresponds to the foliage of a single tree. However, trees are more likely to grow next to trees of the same species, forming groups that cover significant ground surface represented by many pixels in an aerial photo of the investigated area.

The input of the model accepts patches of size $64 \times 64$ so the image of some forest under examination is partitioned into patches of the specific size. Accordingly, the accompanying tree species' map is partitioned to patches of the same size that correspond to the same forest area. Furthermore, the tree species' maps are converted to binary maps with pixel values of zero or one that indicate whether the specific pixel corresponds to the target species or not. This approach creates a dataset comprised from several thousand patches and their corresponding maps. We applied no special preproccessing to the input patches but we exclude patches that contained very few or no trees at all. More specifically, we detected little or no representation of trees by examining the total brightness of each patch since ground texture tends to be much brighter than foliage. An appropriate threshold for deciding which patches to discard because of reduced trees depiction was determined after visual inspection.

### 3.4 Deep learning modelling

Both image segmentation and tree identification make use of an input image and an accompanying pixel-wise map that holds the labels of the represented trees. For image segmentation, labels correspond to image objects and subjects while, for tree identification, labels correspond to tree species.

**Model selection:** The most popular state-of-the-art models for image segmentation are the fully convolutional Dense-Net [27], the multi-scale context aggregation by dilated convolutions [28], the DeepLab model that uses spatial pyramid pooling [29], the FastNet [30] and the U-Net model [31]. The latter was used in biomedical image segmentation and yielded precise results in tasks where getting a class label for each pixel is crucial. Despite its effectiveness, U-Net is very straight-forward to implement and does not require extreme fine-tuning and task-specific architectural modifications. Given the fact that we deal with tasks comprising from binary labels (since we try to distinguish between two tree species at a time), U-Net architecture is considered sufficient for tackling the problem. The name of the U-Net model comes from its shape which is formed by its two data flow paths: the contrastive and the expansive paths as shown in Fig. 3. Input goes through the contrastive path consisting of subsequent convolutional and down-sampling operations until it is considerably reduced in size at the center of the model. After that point, feature maps are processed by subsequent convolutional and up-sampling operations until the size of the labeled maps is reached.

**Design decisions:** The predicted output and the actual output (labeled map) contribute to the loss function of the model, which is cross-entropy. All model layers use the ReLU activation function [32] except the final output layer that uses the sigmoid function, which serves the requirement that every pixel in the output map has a probability of belonging to a certain label that is independent from any other pixel output.

**Avoiding overfitting:** An important architectural characteristic we used to avoid overfitting, caused by the relatively small dataset and the high capacity of the model (consisting of *9,075,201* trainable parameters), is the two dropout [33] layers in the model's contrastive path. These layers randomly drop some elements of the feature maps and prevent co-adaptation of the parameters during training, enhancing the generalization of the model and avoiding overfitting. The model is trained on mini-batches of image patches as input and their maps of tree species identification as targets.

**Dataset imbalance:** Invasive tree species are much less in number than the dominant species of a forest. In the studied cases, invasive trees occupy less than *10%* of the forests' surfaces. This dataset imbalance greatly reflects on the model performance if no countermeasures are applied: the majority class overwelmes the minority class (the invasive species) and the identification of the latter is extremely poor. To this direction, the loss function is weighted appropriately so that the minority class misclassification is penalized to the extend that the dataset imbalance effect cancels out. The loss function is weighted analogous to the number of the invasive trees in a patch: more invasive trees present in a patch translates to a higher misclassification penalty. The calculation for the loss weighting factor $f_M$ of a invasive tree map $M$ is shown below.

$$f_M = \frac{\sum_{i,j} M[i,j]}{a} + 1 \;, \quad M[i,j] = \begin{cases} 1 & \text{if } M[i,j] \text{ represents an invasive tree} \\ 0 & \text{otherwise} \end{cases}$$

The value of $a$ is determined using the ratio of the total number of invasive trees in the dataset over the total number of trees not considered invasive. This ratio is multiplied by the number of pixels in each tree map ($64 \times 64 = 4096$) to provide the value of $a$. Since this ratio is about *10%* in the available datasets, we use $a = 400$.

**Data augmentation:** Image augmentation is also applied to enhance the performance of the model. Since the labels for the invasive tree identification problems are provided in a 2-D species map, any spatial transformation applied to the input patches must also be applied to the target map. For example, an image rotation of the input patch by definition requires an image rotation of the species map in order to maintain the one-to-one correspondence between the input image pixels and the pixels of the output map. To avoid practical problems due to interpolation schemes on the species maps, like distorting their binary values, we only use spatial transformations that do not require pixel interpolation. Such augmentations are image rotations by angles that are multiples of *90* degrees and vertical or horizontal image flipping. We also use four other augmentation schemes based on non-spatial transformations: image sharpness, brightness and color adjustments and random color channel shifting by a small amount.

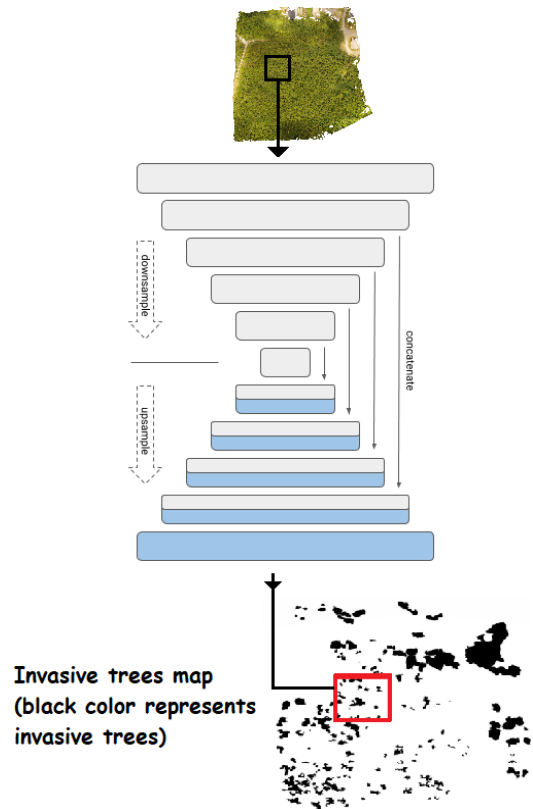

**Fig 3**. The invasive tree identification model with a U-Net architecture. Patches of the orthomosaic were inputted in the model reduced in the size while passing the contrastive path and outputted after resizing and labelling as invasive trees map.

**Performance metrics:** The model performance is measured by simple statistics like false and true positive/negative classification ratios. We particularly measure the percentage of output map pixels that are classified as an invasive species while actually being one (True positive) and the percentage of output map pixels that are classified as invasive species but actually are not (False Positive). These metrics are more appropriate than plain classification success rate because of the class imbalance. The decision threshold of the output neurons is also shifted upwards in order to obtain a good compromise among the performance metrics.

**Data split for testing:** Tuning of the output threshold was performed using a validation set. This essentially means that the dataset is split into a training set comprised from *60%* of the available data, a validation set comprised from *20%* of the data and a test set containing the remaining data. During the creation of the various sets, special care was taken to make sure that all sets contain similar number of samples in terms of invasive species surface coverage. Spefically, the invasive tree percentage coverage of the tree species maps was noted for every patch in the dataset and was categorized as *0%* invasive tree coverage, invasive tree coverage between *1-20%* , *21-50%, 51-80%* and invasive tree coverage in the range *81-100%*. During the split of the patches to the three sets, an equal ratio of the coverage categories was represented in each set. So, the training set, validation set and test set hold an equal percentage of the invasive tree coverage categories. We found that following this balancing approach instead of an n-fold validation approach significantly improved model generalization. The model is trained with the stochastic gradient descent algorihtm with a learning rate of $1 \times 10^{-3}$.

## 4  Results

We examined two cases of tree species identification. The first deals with detection of black locust (site 1 as described in Section 3.1); the second with detection of larch trees (site 2, described in Section 3.1). The validation process determined *0.85* to be an appropriate output threshold value for both cases. Results are shown in Table 1.

**Table 1**. Results of the U-Net model on the black locust and the larch identification problem. True Positive and True Negative values represent the quality of the results of the two identification case studies, black locust and larch trees.

|                | **Black Locust** | **Larch** |
|---|---|---|
| True Positive  | 62.6% | 37.4% |
| True Negative  | 98.1% | 97.7% |

Larch identification has a significantly lower success rate because the corresponding dataset is much smaller than the dataset showing black locust trees. The results are greatly affected by the threshold value of the output layer. Raising the threshold value reduces the correct classification rate of the target tree species and improves the correct classification rate of the non-target trees. For a better model evaluation, we also examine the effect of the invasive trees' proportion within a specific image patch on the classification of at least the number of invasive trees that corresponds to the imbalance ratio between the tree species ($\approx 10\%$). In other words, we observe the rate at which the model identifies at least *10%* of the invasive trees in a patch, in relation to the percentage of the visible invasive trees in the patch. The black locust test set patches are divided into five categories according to the surface covered by the containing invasive trees: zero invasive trees; less than *20%* of the total pixels of the $64 \times 64$ patch; invasive trees covering *20-49%* of the total surface; *50-79%* surface coverage; and *80-100%* coverage. For each patch that the model identifies at least *10%* of invasive trees contained in it, we consider it as being classified correctly. Table 2 shows the results of this experiment.

**Table 2.** Identification percentage of at least 10% of the contained invasive trees in the test set patches in relevance to the percentage of pixels occupied by invasive trees.

| Invasive Trees Coverage (%) | 0 | 1-19 | 20-49 | 50-79 | 80-100 |
|---|---|---|---|---|---|
| Detection of at least 10% of contained invasive trees (%) | 97,04 | 18.88 | 27.54 | 48.8 | 61.7 |

As expected, the blacker locust trees in a patch, the better chance for the model to detect at least the portion of them that corresponds to their statistical frequency of occurrence. Invasive tree clusters triggered the detection of appropriate features more often than scattered invasive trees.

## 5 Discussion

The methodology was successfully applied to our datasets, even though there is much space for improvements. We acknowledge the low detection rates, which mainly related to the insufficient amount and balance of the training data. The results of the first dataset indicated that the amount of data for black locusts is not enough to train a deep learning network. Moreover, the data are imbalanced regarding the classes black pine and black locust, which can be improved by increasing the dataset. In Japan, black locust trees appear mixed with other tree species which makes it difficult to get images from pure black locust stands. Nevertheless, since the invasion of the tree species is a problem in forests all over Japan (e.g. [34], [35]) images of black locust in mixed forests can be easily collected. Our second dataset dealt with images of the mixed forest in YURF. We ran the deep learning model for only one of the orthomosaics to get a first idea of the precision we could achieve. We assume that the accuracy of the deep learning application can be increased by using data from the whole season (spring, summer and autumn).

### 5.1 Study implications

Coastal forests in Japan have peculiar stand conditions (sandy soils, high salinities and strong winds) and both black pine and black locust tree species are adapted to these conditions. The invasion had changed the monoculture plantation into a mixed forest, which might lead to an enhancement of forest resilience, since several studies have pointed out the benefits of mixed forests under climate change. Tinya et al. [36] suggested the increase of stability in forest stands to stress and disturbances that can be mitigated by mixed forests. The image analysis performed offers the opportunity to study the benefits of the forest mixture and helps to characterize the resilience of this mixed forest. Even though the black locust can have positive effects on the costal forest, negative influences are discussed as well. Since black locust trees are deciduous, they will provide wind tunnels in the leaf absent season. We assume that the distribution of the trees has a significant influence on the windbreak potential of the coastal forests. On the other hand, larch trees, which were studied via our second dataset, are part of a different kind of mixed forest. The focus of previous studies has been to evaluate processes in forests in relation to the tree species [37] or the composition of the forest stands [37], [38]. These studies tried to solve important aspects of forests and propose solutions for mitigating the effects of climate change, focusing though only on small forest patches. Thus, a reliable methodology for scaling up to forest stands can provide insights not only about tree species composition but also contribute to understand other essential forest characteristics (soil type, soil moisture, fungi, nutrient cycles, etc.). The classification of the mixed forest via computer vision, such as the work in this study, is important to further achieve these goals.

### 5.2 Policy-making

The methodology proposed in this paper is a tool for forest management practices since it can provide fast and reliable information about forests. In particular, the coastal forests with their functions as wind and tsunami breakers could benefit from a more automated management system. Further, the Ministry of Environment in Japan has called for the urgency of management issues of black locust species [35]. Since most of the forests are unmanaged and dense, our study provides a simple tool for forestry to assess the spread of black locusts and provides the possibility to detect invasive trees. Furthermore, since mountainous forests are steep and hard to access, the methodology of this study partially solves the problem of inaccessibility. Since the technique uses merely images, this methodology can be applied for forestry classification/management world-wide, contributing to more focalized field surveys.

### 5.3 Limitations and assumptions

Our study shows that we are able to identify tree species, however, we also face some limitations. In the first case study, the orthomosaic used showed less than 10% of black locust trees. Fieldwork in this area showed that the amount of black locust is higher than the 10%, as shown in the orthomosaic, which can be explained by their smaller height (12-18 m) in comparison to black pine trees (up to 40 m). The smaller black locusts are often covered by the black pine canopies but they are partly still visible since black locust trees mainly appear in gaps formed in between the black pines. The true distribution and number of trees is still unknown, since not all trees are visible on the images. Therefore, further fieldwork needs to be conducted to evaluate the maps generated by our model. Even though the method has its limitations it provides a fast overview of the study area and facilitates management approaches. In the second case, we attempted to identify larch trees in a mixed forest, where they are one of the dominant tree species. Larch tree structure makes it easier to recognize them from the images. The general idea of this approach was to see how well the deep learning network can deal with these images since a further step will be the classification of all dominant tree species in the mixed forest. Field surveys indicated that in our study areas there are several trees which are covered by canopies of taller trees. Thus, we acknowledge that this methodology might not get all the trees in the forest and may only provide information of the visible and dominant tree species. Since in situ fieldwork is barely possible in these areas and knowledge gaps about mixed forests are still large [24], our methodology can be considered as a helpful tool for forestry research.

### 5.4 Future work

For future work, we aim to overcome some of the limitations mentioned by using satellite data to increase the amount of data used as input to our models. We plan to locate regions in North America where black locust is a native species and acquire satellite photos to augment our dataset and reduce the existing imbalance of the data. We also plan to acquire images of different seasons (spring, summer and autumn), which will help to increase the accuracy of the model, something already shown as a basic solution in related work [39]. In this context, the work in [40] demonstrated the effectiveness of a multi-temporal dataset on image classification issues. Generating synthetic data is another option worth considering [41], [42].

A comparison of satellite and drone images for tree species' classification for deep learning applications is an interesting aspect we also plan to work on, primarily for the black locust problem. Prior fieldwork results in the larch-oak mixed forest site showed that our study area includes more than 20 different tree species, although some of them are in small numbers. Therefore, we will focus on the dominant tree species of the mixed forest, namely by increasing the training data for larch, oak and beech trees.

Further approaches, for instance the additional use of spatial information along with the images are not considered, since natural forest structures, tree species' compositions and the behavior of the invasive tree species are irregular and not well-known.

## 6   Conclusion

Our study aimed to identify two kinds of tree species in Japanese mixed forest by using UAV-acquired RGB images and deep learning technologies. Our results indicated that it is possible to classify black locust and larch trees by using these two technologies. The model was able to identify patches without black locust/larch with high accuracies but showed lower accuracies for detecting the target species. The main reason was the imbalance in our input data. The number of images representing black locust and larch trees was significantly lower than the other trees. Even though our data were highly imbalanced, the results are promising for future work in this field, since our proposed methodology is suitable for large-scale forestry management applications. Further data acquisition efforts are planned for increasing our datasets and improve the performance of our model.

### Acknowledgments

Andreas Kamilaris and Savvas Karatsiolis have received funding from the European Union's Horizon 2020 research and innovation programme under grant agreement No 739578 complemented by the Government of the Republic of Cyprus through the Directorate General for European Programmes, Coordination and Development.